\documentclass[letterpaper, 10 pt, conference]{ieeeconf}  

\IEEEoverridecommandlockouts                              

\usepackage{graphics} 
\usepackage{epsfig} 
\usepackage{mathptmx} 
\usepackage{times} 
\usepackage{amsmath} 
\usepackage{amssymb}  
\usepackage{comment}
\usepackage[T1]{fontenc}
\usepackage{multirow}
\usepackage{textcomp}
\usepackage{gensymb}
\usepackage{balance}
\usepackage{soul}
\usepackage{color}
\graphicspath{{images/}}
\usepackage{adjustbox}
\usepackage[utf8]{inputenc}
\usepackage{csquotes}
\usepackage[english]{babel}
\usepackage[hidelinks]{hyperref}
\usepackage[euler]{textgreek}
\usepackage{bm}

\usepackage{graphicx}
\graphicspath{ {./images_new/} }

\usepackage{bm}

\usepackage{float} 
\usepackage{caption}
\usepackage{subcaption}

\usepackage[backend=biber,style=numeric,sorting=none, maxnames=99, giveninits=true]{biblatex} 
\title{\LARGE \bf
Shakebot: A Low-cost, Open-source Robotic Shake Table for Earthquake Research and Education}

\author{Zhiang Chen, Devin Keating, Yash Shethwala, Aravind Adhith Pandian Saravanakumaran, \\ Ramon Arrowsmith, Albert Kottke, Christine Wittich, Jnaneshwar Das
}

\begin{document}
\renewcommand{\labelenumi}{\arabic{enumi}.}
\renewcommand{\labelenumii}{\arabic{enumi}.\arabic{enumii}}
\renewcommand{\labelenumiii}{\arabic{enumi}.\arabic{enumii}.\arabic{enumiii}}
\renewcommand{\labelenumiv}{\arabic{enumi}.\arabic{enumii}.\arabic{enumiii}.\arabic{enumiv}}

\maketitle
\thispagestyle{empty}
\pagestyle{empty}
\maxdeadcycles=200

\begin{abstract}
Shake tables serve as a critical tool for simulating earthquake events and testing the response of structures to seismic forces. However, existing shake tables are either expensive or proprietary. This paper presents the design and implementation of a low-cost, open-source shake table named \textit{Shakebot} for earthquake engineering research and education, built using Robot Operating System (ROS) and principles of robotics. The Shakebot adapts affordable and high-accuracy components from 3D printers, particularly a closed-loop stepper motor for actuation and a toothed belt for transmission. The stepper motor enables the bed to reach a maximum horizontal acceleration of 11.8 $m/s^2$ (1.2 $\mathbf{g}$), and velocity of 0.5 $m/s$, with a 2 $kg$ specimen. The Shakebot is equipped with an accelerometer and a high frame-rate camera for bed motion estimation. The low cost and easy use make the Shakebot accessible to a wide range of users, including students, educators, and researchers in resource-constrained settings. 

The Shakebot, along with its digital twin--a virtual shake robot--has showcased significant potential in advancing ground motion research. Specifically, this study examines the dynamics of precariously balanced rocks. The Shakebot provides an approach to validate the simulation through physical experiments. The ROS-based perception and motion software facilitates the code transition from our virtual shake robot to the physical Shakebot. The reuse of the control programs ensures that the implemented ground motions are consistent for both the simulation and physical experiments, which is critical to validate our simulation experiments. 
\end{abstract}

\section{Introduction}
For decades, researchers have used shake tables to simulate the effects of earthquakes on structures, systems, and materials, which provides a critical approach to earthquake research and education~\cite{jacobsen1929vibration, lu2008shake}. Simulated earthquake conditions can help civil engineers identify potential weaknesses in the design and suggest ways to improve the structure's resilience~\cite{bogdanovic2019shake, rakicevic2021hybrid}. In a controlled laboratory setting, shake tables can also be used to study the behavior of fragile geological features, which aids seismologists in finding upper-bound constraints on ground motions~\cite{purvance2008freestanding}. By simulating earthquakes in the classroom, students can better understand the forces involved and the factors that influence the behavior of structures~\cite{elgamal2005line}, inspiring the next generation of engineers and seismologists.

\begin{figure}[h]
\captionsetup[subfigure]{justification=centering}
\begin{subfigure}[t]{0.48\textwidth}
    \includegraphics[width=\textwidth]{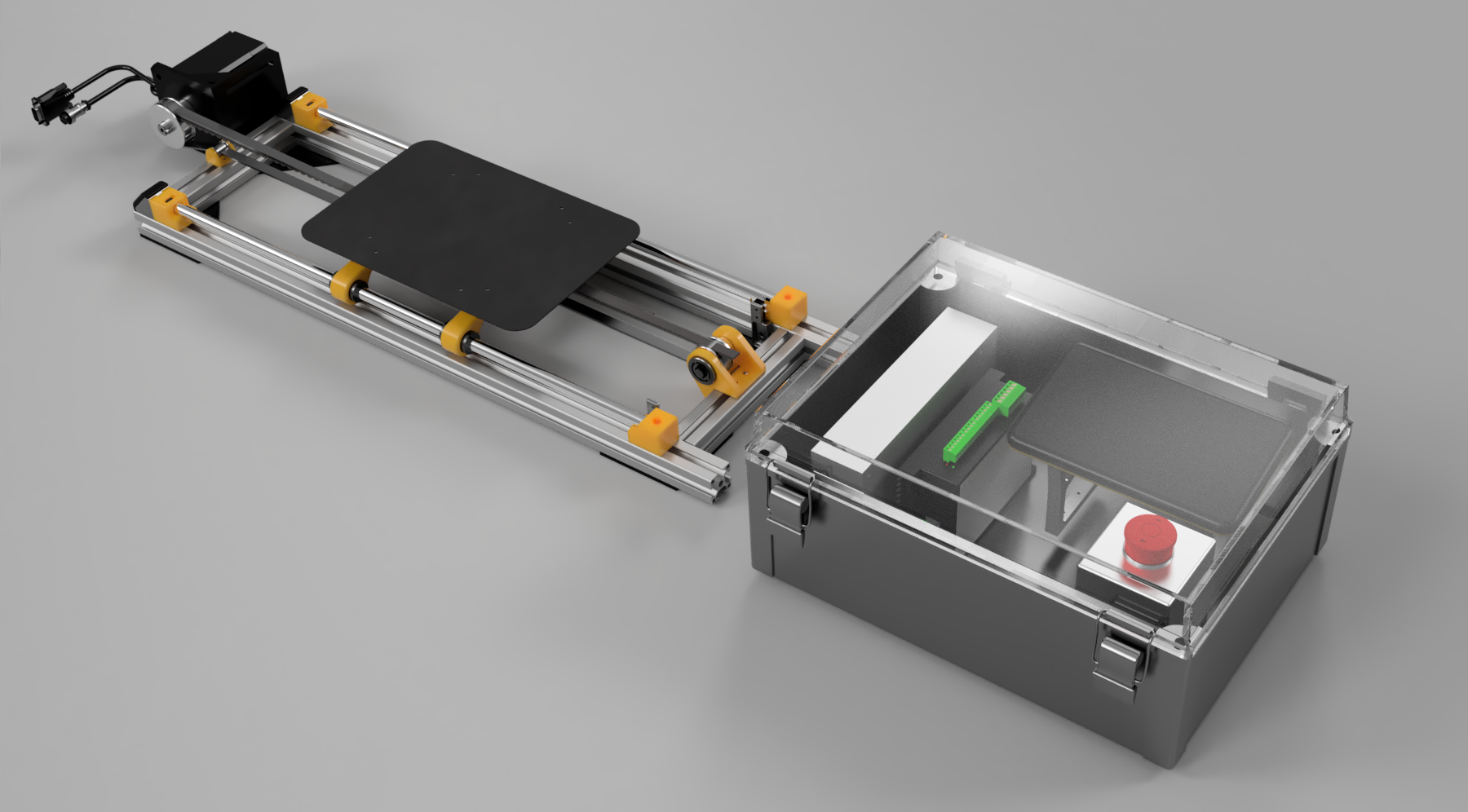}
    \caption{Shakebot CAD model}
    \label{fig:shakebot_CADModel}
\end{subfigure}
\begin{subfigure}[t]{0.48\textwidth}
    \centering
    \includegraphics[width=\textwidth]{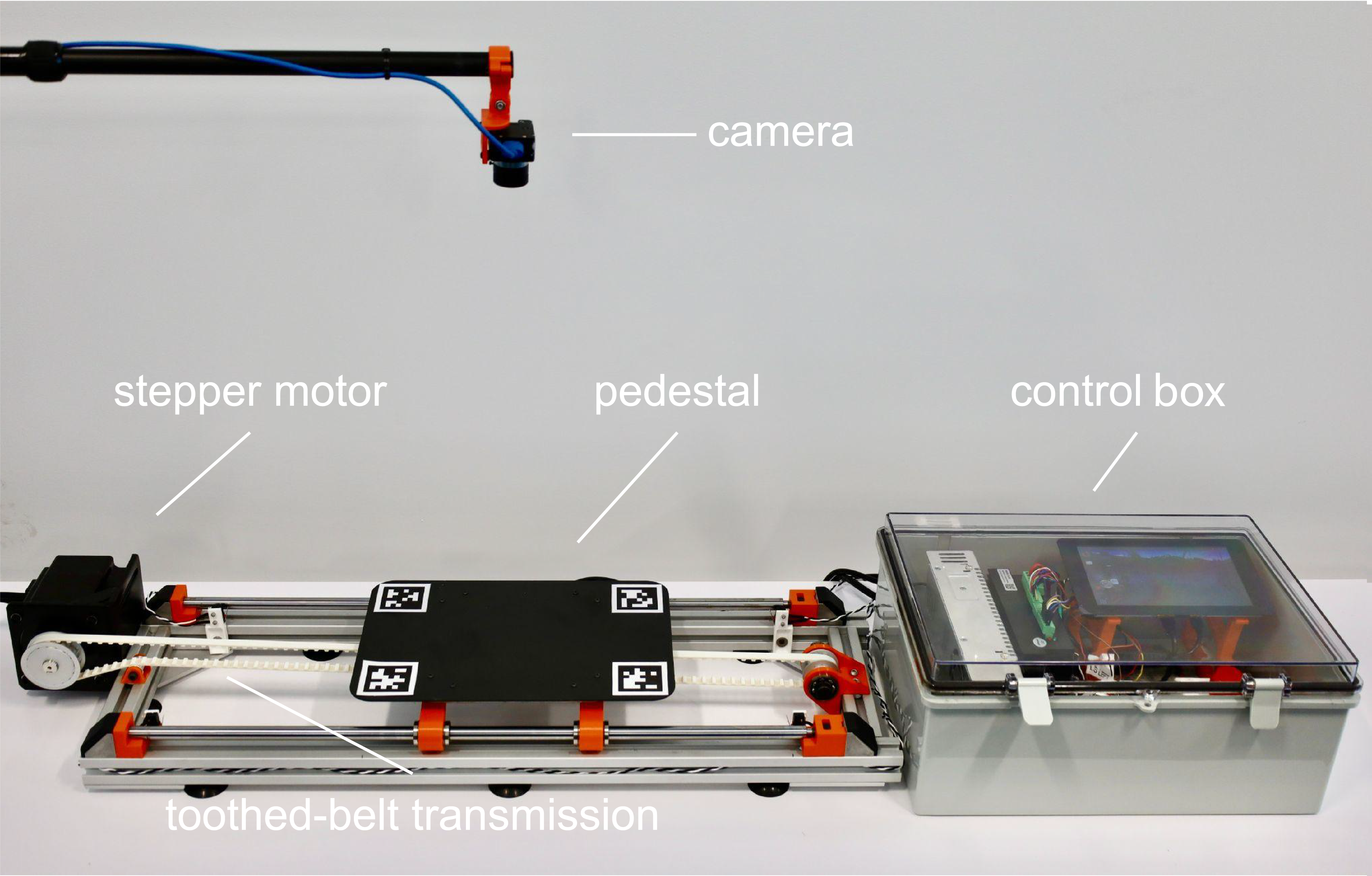}
    \caption{Shakebot experimental setup}
    \label{fig:shakebot_setup}
\end{subfigure}
\caption{Shakebot CAD model and actual experimental setup. Stepper motor actuates pedestal through toothed belt and pulley transmission. Control box includes power supply, stepper motor driver, Raspberry Pi, emergency push button, and touchscreen. For scale, the control box has a dimension of 390$\times$290$\times$160 $mm$.}
\label{fig:shakebot}
\end{figure}

An important application of shake tables is to examine the dynamics of precariously balanced rocks (PBRs), which are free-standing rocks balanced on, but not attached to sub-horizontal pedestals~\cite{brune1996precariously, shi1996rocking, anooshehpoor2004methodology}. Because an earthquake ground motion of sufficient amplitude and duration may topple PBRs, the fragile configurations of PBRs provide upper-bound constraints on ground motions, refining probabilistic seismic hazard analysis~\cite{anderson2014precarious, rood2020earthquake}. The dynamics of the PBR overturning process are non-linear \cite{yim_rocking_1980, purvance2008freestanding}. Purvance et al. \cite{purvance2008freestanding} used a shake table to study the PBR overturning dynamics. However, difficulties can arise when using conventional shake tables to repeat overturning experiments. For example, when analyzing PBR fragility, shake experiments would involve accurate and repeated positioning of a large boulder on the shake table to record the overturning responses given ground motions. These boulders are often so heavy that the use of a crane or forklift is needed, and repositioning them precisely is difficult (e.g., \cite{saifullah2022seismic}), complicating experiment repeatability.

\begin{table*}[htpb]
\vspace{8pt}
\caption{Shake tables and their specifications.}
\begin{tabular}{ccccccccc}
\hline
                    & Open source & Cost (USD) & Payload ($kg$) & Vel ($m/s$) & Acc ($\mathbf{g}$) & Software & Programmable & Transmission        \\ \hline
Quanser \cite{web:qanser_st2}            & $\times$  & 30k         & 7.5          & 0.4     & 2.5     & NA       & \checkmark            & ball screw          \\
CUSHAKE  \cite{baran2011construction}           & $\times$  & 45k      & 3500         & 0.01      & 1       & DEPSIM   & \checkmark            & threaded rod        \\
Kelvin  \cite{web:kelvin_es}           & $\times$  & 1.3k       & NA           & NA        & NA      & NA       & $\times$            & reciprocating arm   \\
SARSAR   \cite{damci2019}           & \checkmark  & 1.8k        & 200 (0.4 $\mathbf{g}$)         & 0.35      & 1.5     & NA       & \checkmark            & ball screw          \\
Shao and Enyart  \cite{shao2014development}   & $\times$  & NA         & 228          & NA        & 4       & NA       & \checkmark            & hydraulic actuators \\
Kınay    \cite{kinay2006construction}           & \checkmark  & NA         & 80           & 0.25      & 2       & Scilab   & \checkmark            & ball screw          \\
QuakeLogic \cite{web:testbox_shaketable} & $\times$  & 15k         & 50           & 0.5       & 1       & TESTLAB  & \checkmark            & ball screw          \\
Shakebot (ours)           & \checkmark  & 1.3k       & 2            & 0.5       & 1.2     & ROS      & \checkmark            & toothed belt        \\ \hline
\multicolumn{9}{c}{\begin{tabular}[c]{@{}l@{}}
Note: Shakebot cost does not include camera or logistics. The cost of SARSAR is estimated based on available items in the United States. \\ The Specifications of payload, velocity, and acceleration are the maximum values. SARSAR has 200 $kg$ maximum payload at 0.4 $\mathbf{g}$ acceleration. 
\end{tabular}}
\end{tabular}
\label{tab:comparison_table}
\end{table*}

In an effort to reduce variability, explore a broad parameter space of ground motions, and automate PBR overturning analyses, our previous study developed a virtual shake robot in simulation \cite{chen2024virtual}, which is a digital twin of the Shakebot. The virtual shake robot utilized core technologies in robotics, such as Robot Operating System (ROS), Gazebo simulation toolbox, and Bullet physics engine. Although the virtual shake robot allows us to conduct overturning experiments in simulation, the results need to be validated through physical experiments. To complete this validation, we designed and built a low-cost, small-scale shake table, called \textit{Shakebot} (Fig.~\ref{fig:shakebot}). 

The Shakebot provides a reverse method for simulation validation. Physics engines have been used for robotic development. However, when robotic systems are transitioned from simulations to the real world, simulation gaps may cause critical failures (e.g., in the arena of reinforcement learning~\cite{zhao2020sim, ding2020challenges}). When using a physics engine for scientific studies, we also need to address the simulation gap to calibrate and scale-up the simulation experiments. Using the Shakebot, we test the overturning dynamics of down-scaled, 3D-printed PBRs. We build the same PBR models in simulation and validate the overturning dynamics using our virtual shake robot. Following this approach, we are able to compare the overturning patterns obtained from the Shakebot and the virtual shake robot, further allowing us to quantify the uncertainty in simulation. The Shakebot serves as a validation tool for simulation studies, aiding in learning more about the real-time dynamics of the PBRs and potentially calibrating physics engine parameters to reduce the simulation gaps.

The Shakebot demonstrates interdisciplinary research in the use of automated robotic systems for geoscience studies, and enhances the impact of robotics in the natural sciences. Besides its contribution to seismology, the key contributions of this study include: \
\begin{itemize}
\item Shakebot is the first tabletop shake table developed based on robotic concepts (perception and motion systems) and tools (ROS). Because of using ROS, the control systems are reused from our previous virtual shake robot, ensuring that the ground motions (the motion of the bed) are consistent in physical and simulation experiments. This consistency is vital for simulation validation. 
\item The hardware mechanism is simplified by design. The Shakebot adopts a closed-loop stepper motor with a toothed belt and pulley transmission mechanism. Most of the mechanical parts and components are off-the-shelf, and only a few parts are 3D-printed. The simplified design lowers costs and expedites prototyping. Additionally, for safety anticipation, we employ a double-emergency mechanism to allow system override in case of malfunction. 
\item Compared with traditional tabletop shake tables that only use accelerometers, our Shakebot also leverages a top-down view camera to estimate ground motion. The perception system provides an option of fusing accelerations and camera-based displacements for ground velocity estimation. 
\item For the perception system, we have developed a calibration method that samples measurements to construct an overdetermined system. This system is then solved using the Moore-Penrose inverse, enabling the precise estimation of the calibration parameters.
\item We have open-sourced our hardware and software to promote robotic applications in seismic research and education: \url{https://github.com/DREAMS-lab/asu_shake_table}. Despite focusing on PBR dynamics in this study, the Shakebot provides an affordable and accessible platform for other earthquake research and education.
\end{itemize}

The paper is structured as follows. In Sec.~\ref{sec:related-work}, we discuss commercial and academic shake tables available for civil engineering and seismology, and highlight a few of their capabilities and limitations. From Sec.~\ref{sec:sys-desc} to Sec.~\ref{sec:calibration}, we describe the mechanical design, the perception system, the motion system, and the calibration system. In Sec.~\ref{sec:experiment}, we discuss the results of our experiments, followed by our conclusions and future directions in Sec.~\ref{sec:conclusion}.

\section{Related Work}
\label{sec:related-work}
Full- or large-scale shake tables and tabletop shake tables have different applications for research and education. Full-scale shake tables are designed to simulate earthquakes on full-scale buildings, bridges, and other large structures \cite{gavridou2017shake}. Full-scale shake tables, which usually incorporate acceleration and displacement sensors for ground motion estimations, are expensive to build and maintain. The high cost limits their availability in research and education. Full-scale shake tables may also involve safety considerations in operation and damage prevention. 

Tabletop shake tables, on the other hand, are typically used to test smaller models and structure components. Because of their low cost, these tables are also useful in educational settings to teach students about the behavior of structures under earthquake conditions. TABLE \ref{tab:comparison_table} compares existing tabletop shake tables and the Shakebot. Existing tabletop shake tables are comparatively expensive \cite{web:qanser_st2, baran2011construction, web:testbox_shaketable} or only allow a fixed ground motion pattern \cite{web:kelvin_es}. None of the existing tabletop shake tables have open-sourced their hardware. Another limitation of existing shake tables is that they solely employ accelerometers to measure ground motions.

\section{System Description}
\label{sec:sys-desc}
In this section, we discuss the mechanical design, motor selection strategy, and safety mechanism. The system is designed to satisfy the ground motion requirements---maximum dynamic acceleration of 11.8 $m/s^2$ (1.2 $\mathbf{g}$), maximum velocity of 0.5 $m/s$, and maximum displacement of 0.45 $m$, with maximum specimen (e.g., PBR) payload mass of 2 $kg$. 

\subsection{Hardware Mechanism}
The two main components of the Shakebot are a chassis and a control box, as shown in Fig.~\ref{fig:shakebot}. The chassis is constructed of an extruded aluminum T-slot frame that supports two linear shafts (625 $mm$ length for each shaft). A carriage is mounted on the linear shafts with ball bearings, which reduce the friction between the linear shafts and the carriage. A flatbed (pedestal) is attached to the carriage to hold specimens. A stepper motor (i.e., NEMA 34) actuates the carriage through a toothed belt and pulley transmission. The stepper motor driver (CL86T), contained in the control box, drives the stepper motor. The control box also houses the power supply (S-350-60), an emergency stop button, a touchscreen for the user interface, and a Raspberry Pi that handles trajectory generation and low-level system control.

\subsection{Motor Selection}
To select a stepper motor, we considered the stepper motor torque at its maximum angular velocity to satisfy the required ground motion acceleration and velocity. Because stepper motor torque decreases with velocity, if the torque at the maximum velocity satisfies the acceleration requirement, the stepper motor can also achieve so at any smaller velocities. We selected Nema 34HS31, a closed-loop stepper motor (with an encoder) with a torque output of 1.56 $Nm$ at the maximum speed of 1200 RPM. The maximum translational force is calculated from the torque, 
    \begin{equation}\label{eqn:motor_eqn}
    \centering
        \begin{aligned}
        F = ma = \frac{\tau}{r}
        \end{aligned}
    \end{equation}
where $F$ is the translational force, $r$ is the outside radius of the toothed pulley, $m$ is the maximum payload mass (including carriage and specimen), $a$ is the maximum translational acceleration, and $\tau$ is the stepper motor torque. We select a toothed pulley with an outside radius of 25.91 $mm$. The translational force at the maximum speed is 60.21 $N$. Considering a maximum total payload mass $m$ of 4 $kg$ (including a 2 $kg$ carriage and 2 $kg$ specimen), the corresponding acceleration produced from the translational motion is 15.1 $m/s^2$ (1.54 $\mathbf{g}$). The ratio of the calculated acceleration to the required acceleration is 1.28, which is adequate empirically to compensate for the friction and energy losses in transmission. Additionally, at the maximum stepper motor speed of 1200 RPM, the translational speed can reach 0.52 $m/s$, which also satisfies our design requirement for PBR studies. Note that we used the outside radius on the specifications of the toothed pulley for the calculation here. The outside radius value is appropriate for the stepper motor selection. We conducted a calibration process that uses a camera and several fiducial markers to estimate this radius precisely for accurate control. The calibration process is discussed in Sec.~\ref{sec:velocity_controller_calibration}. 

The stepper motor driver reads signals from the encoder attached to the stepper motor, forming a closed-loop system to control the stepper motor. The microstepping technology in the stepper motor driver allows a higher step resolution (1.8°/2000 per step), which achieves smooth motions at low speeds (as low as 0.001 $m/s$ in our implementation).

\subsection{Safety Mechanism}
\begin{figure}[h]
        \centering
        \includegraphics[width=0.48\textwidth]{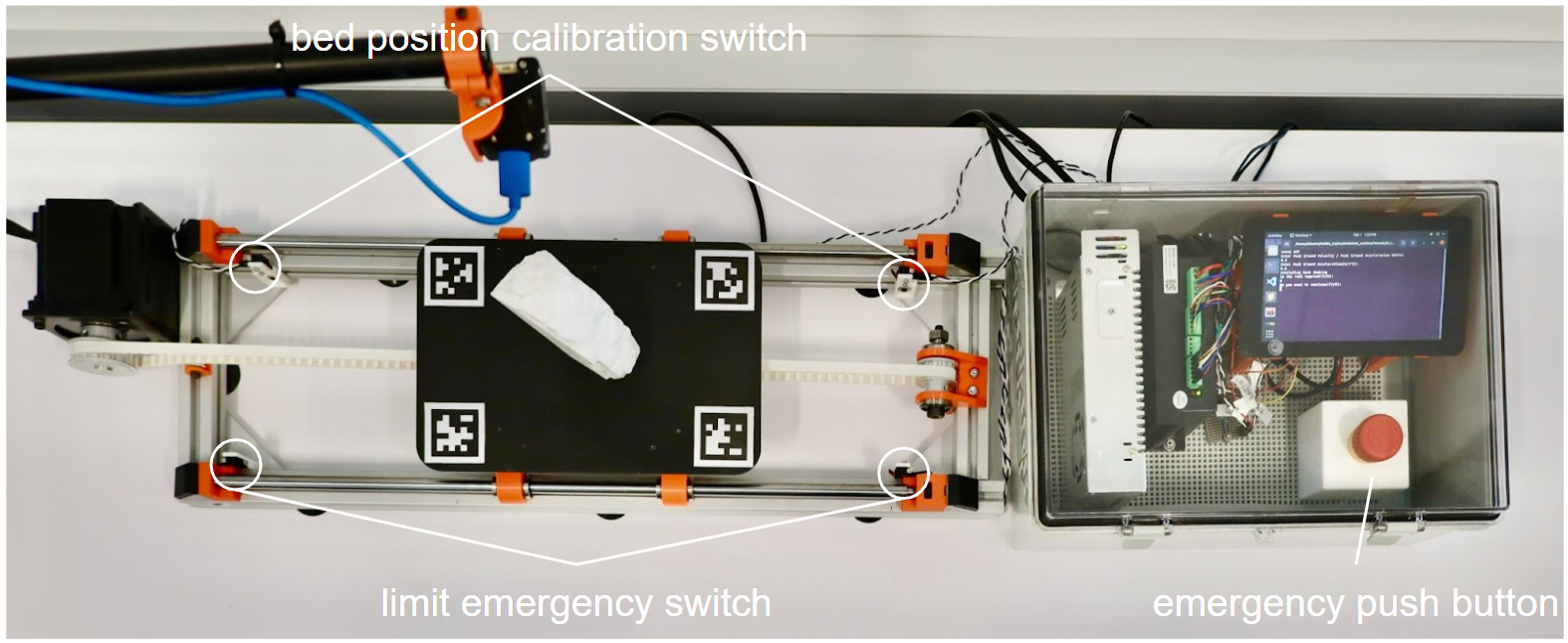}
        \caption{Shakebot from top-down view illustrating double emergency mechanism. }
        \label{fig:top-down_Shakebot}
    \end{figure}

The design incorporates two levels of safety features. The first level consists of a pair of limit emergency switches situated at either end of the carriage's range of motion, as shown in Fig. \ref{fig:top-down_Shakebot}. These switches link to the emergency stop function on the stepper motor driver. Once either of these switches is triggered, the stepper motor is disabled. The second level of protection is supplied by an emergency push button in the control box, which disconnects the power supply from the stepper motor driver. The emergency push button can be manually activated by the Shakebot's operator. Fig. \ref{fig:top-down_Shakebot} shows a top-down view of the Shakebot.  

\section{Perception System}
\label{sec:perception-system}
The perception system estimates the status of the bed using a camera and an accelerometer. Four fiducial markers are affixed to the corners of the bed (Fig.~\ref{fig:shakebot_setup} and Fig.~\ref{fig:top-down_Shakebot}). A top-down view camera (FLIR Chameleon\textsuperscript{®}3 Color Camera CM3-U3-13Y3C-CS 1/2") captures the position and orientation of the bed using the ROS package \textit{apriltag\_ros} \cite{malyuta2017mt, Wang2016}. Additionally, an accelerometer (Wit Motion HWT905-TTL) attached underneath the bed measures acceleration (ground motion acceleration). We directly obtain ground motion displacement and acceleration from the fiducial marker detection and accelerometer, respectively. We additionally estimate the ground motion velocities by fusing the information from the fiducial marker detection and accelerometer. 

\subsection{Displacement Estimation} 
\label{position_recorder}
The camera estimates ground motion displacement by detecting the fiducial markers at each corner of the bed. The \textit{apriltag\_ros} package provides the pose of each visible marker in the camera coordinates. When the bed moves from one position to another, we obtain the relative pose of a marker attached to the bed, 
\begin{equation} \label{eqn:settinghome}
    H_i^j = H_c^j \cdot \  H_i^c = [H_j^c]^{-1} \cdot \  H_i^c   
\end{equation}
where $i$ and $j$ indicate marker positions at two time stamps, $c$ indicates the camera coordinates, $H_a^b$ is the transformation matrix for position $a$ with respect to position $b$. $H_j^c$ and $H_i^c$ are obtained using the \textit{apriltag\_ros} package.

We extract the translation vector from $H_i^j$ as the relative displacement of the fiducial marker at two positions. We estimate ground motion displacement by averaging the relative displacements from all visible fiducial marker detections. When a fiducial marker is obstructed by a specimen, the ground motion displacement estimation only relies upon the remaining visible fiducial markers. Note that the relative displacement from \textit{apriltag\_ros} is inaccurate despite intrinsic camera calibration. To address this issue, perception calibration is performed (Sec.~\ref{sec:perception_calibration}).

\subsection{Acceleration Estimation}
The accelerometer installed beneath the bed measures the ground motion acceleration vector in three dimensions. However, the bed's motion is limited to one degree-of-freedom prismatic movement. To obtain the acceleration along the bed movement direction, we align the accelerometer's x-axis with the bed's direction of motion. However, the manual alignment is not perfect. We additionally calculate the absolute value (Euclidean distance) of the acceleration vector, because the absolute acceleration value is the same value of the acceleration vector along the bed movement direction. To eliminate high-frequency vibration, we apply a low-pass filter to the raw acceleration data. 

\subsection{Velocity Estimation}
\begin{figure}[h]
\vspace{8pt}
    \centering
    \includegraphics[width=0.5\textwidth]{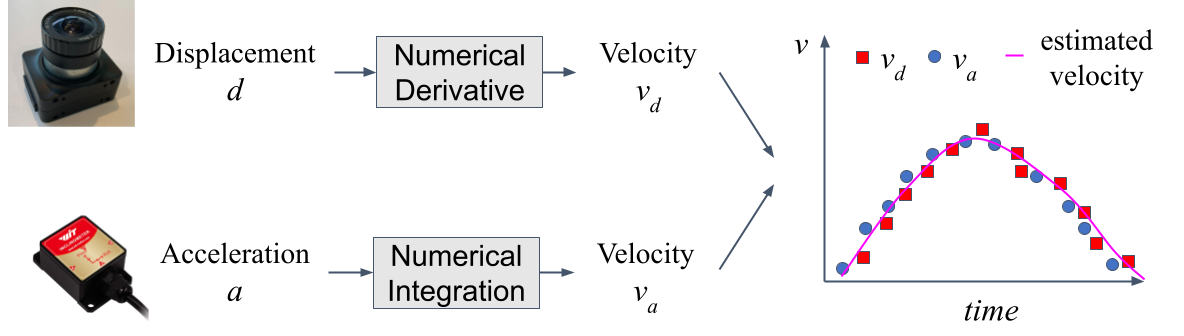}
    \caption{Velocity fusion workflow. Estimated velocity is regression of velocities from displacement and acceleration.}
    \label{fig:velocity_fusion}
\end{figure}

We estimate ground motion velocity by fusing the derived velocities from the fiducial marker detection and accelerometer data. Fig.~\ref{fig:velocity_fusion} illustrates the velocity fusion process. During a shake experiment, the timestamped ground motion displacements and accelerations are recorded. After the shake experiment, we implement numerical derivation to obtain velocity $v_{d}(t)$ from the ground motion displacements d(t).
From the ground motion accelerations $a(t)$, we conduct numerical integration to obtain velocity $v_{a}(t)$.

From the recordings of fiducial markers and acceleremeter, we obtain two sets of derived velocities: $\mathbf{v_d} = \{v_d(t_1),...,v_d(t_M)\}$ and $\mathbf{v_a} = \{v_a(t_1),...,v_a(t_N)\}$. We phrase the ground motion velocity estimation as a regression problem with an objective function that minimizes the square of errors between estimated and derived velocities,
\begin{equation} 
   \arg\min_{\theta} || \hat{v}(t; \theta)-\mathbf{v_s}(t) ||
\end{equation}
where $\hat{v}(t; \theta)$ is a parametric ground motion velocity function, and $\mathbf{v_s}=\mathbf{v_d} \cup \mathbf{v_a}$ is a set of velocities derived from measurements. In this ground motion velocity estimation setting, $\{v_d(t)\}$ and $\{v_a(t)\}$ need not be synchronized, simplifying the electronic hardware configuration of our Shakebot. The fiducial marker detection and accelerometer have different frame rates. The parametric regression function $\hat{v}(t; \theta)$ has many options. We use a 6\textsuperscript{rd} order polynomial function for demonstration in Sec.~\ref{sec:experiment}. 

\section{Motion System}
\label{sec:control-system}
The Shakebot has a hierarchical motion system and a user interface (UI) to facilitate shake experiments. The hierarchical motion system provides two options to emulate ground motions: single-pulse cosine displacement motion and realistic ground motion from seismogram acceleration recordings. The UI assists users in repeating the shake experiment where single-pulse cosine displacement motions are deployed with different ground motion parameters.  

\begin{figure}[h]
\vspace{8pt}
    \centering
    \includegraphics[scale=0.6]{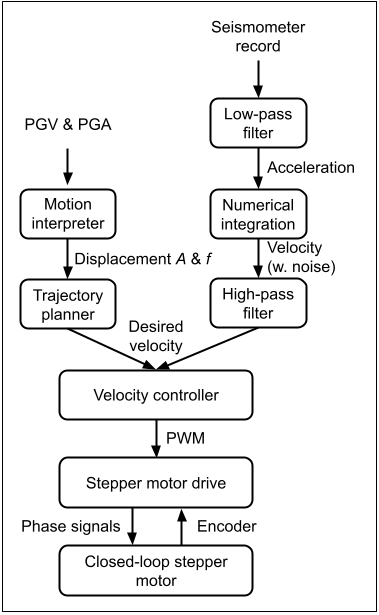}
    \caption{Motion system supports two types of ground motions: single-pulse cosine displacement and realistic ground motion from seismogram acceleration recordings. Single-pulse cosine displacement is defined by peak ground velocity (PGV) and peak ground acceleration (PGA).}
    \label{fig:Hierarchical_Control_System}
\end{figure}

\subsection{Motor Control} 
The Shakebot motion is enabled by a hierarchical motion planner in the Raspberry Pi, which outputs PWM signals to the closed-loop stepper motor driver. Fig.~\ref{fig:Hierarchical_Control_System} shows the hierarchical motion system. The first ground motion option is a single-pulse cosine displacement,
\begin{equation}\label{eqn:displacement}
    d(t) = A - A \cos(2 \pi f t),
\end{equation}
where $d(t)$ is the ground displacement function, $A$ is the amplitude, $f$ is the frequency, and $t \in [ 0 , 1/f]$ is time. $A$ and $f$ are derived from ground motion parameters,
\begin{align}
\label{eqn:conversion}
    &f = \frac{1}{2\pi\kappa} \\
\label{eqn:a_conversion}
    &A = \frac{\alpha g}{4 \pi^2 f^2}  
\end{align}
where $\alpha$ is the peak ground acceleration (PGA), $\kappa$ is the ratio of peak ground velocity to peak ground acceleration (PGV/PGA), and $\mathbf{g}$ is the gravitational acceleration. As shown in Fig.~\ref{fig:Hierarchical_Control_System}, the motion interpreter converts (PGV, PGA) to ($A, f$) using Eqs.~\ref{eqn:conversion}-\ref{eqn:a_conversion}. With ($A, f$), the trajectory planner takes derivative of the cosine displacement function (Eq.~\ref{eqn:displacement}) to obtain ground velocity function, 
\begin{equation}\label{eqn:velocity}
    v(t) = 2 \pi A f \sin(2 \pi f t)
\end{equation}
where $v(t)$ is the ground velocity function, and $t \in [ 0 , 1/f]$ is time. From the ground velocity function, we uniformly sample velocity points $\{v\}$ as the input of the velocity controller. The sampling frequency is a user-defined parameter (usually 200 Hz). Note that the velocity commands $\{v\}$ are desired translational velocities of the bed. The velocity controller converts the translation velocity to the angular velocity of the stepper motor.
The angular velocity commands are further converted to PWM signals for the input of the stepper motor driver. The stepper motor driver, stepper motor, and encoder attached to the stepper motor form a closed-loop control to execute the PWM signals. 

Besides the cosine ground displacement function, the Shakebot
also supports ground motion from observed seismograms. As shown in Fig.~\ref{fig:Hierarchical_Control_System}, we first implement a low-pass filter to remove the high-frequency noise in the raw acceleration data. Numerical integration computes velocities from the accelerations. Then we have a high-pass filter to remove the low-frequency noise in the velocities, because the low-frequency noise in velocities may cause accumulated displacement errors. The output of the high-pass filter is a set of velocities $\{v\}$, which have the same format as the output from the trajectory planner. Because both the trajectory planner and high-pass filter had the same output format, we share a velocity controller to process the desired velocity commands.

\subsection{Control User Interface}
The UI is a portal through which users conduct shake experiments to study PBR overturning responses. 
Once initialized, the UI prompts the user to place a PBR on the bed. Then the user needs to input (PGV/PGA, PGA), which is converted to (PGV, PGA). The motion module (Fig.~\ref{fig:Hierarchical_Control_System}) executes a single-pulse cosine displacement ground motion based on the (PGV, PGA). After the ground motion is completed, the UI prompts the user to input the overturning response of the PBR (i.e., being toppled or remaining balanced). The overturning responses with ground motion parameters and perception estimates are saved in a CSV file. Based on the user's input, the UI exits or continues to conduct a new overturning experiment.  

\section{System Calibration}
\label{sec:calibration}
The Shakebot needs to be calibrated before conducting the shake experiments. We first calibrate the camera intrinsics and fiducial marker detection, and this calibrated perception system is used to facilitate velocity controller calibration. 

\subsection{Perception System Calibration}
\label{sec:perception_calibration}
The perception system requires intrinsic camera calibration and fiducial marker detection scaling. We used the {\textit{camera\_calibration}} ROS package \cite{web:camera_calibration} to obtain the intrinsic camera parameters including scale factor, focal length, pixel dimension, and distortion. Despite the intrinsic camera calibration, the translation estimation from \textit{apriltag\_ros} was found to be inaccurate in our implementation. To calibrate this fiducial marker detection process, we estimate a multiplicative factor between the fiducial marker displacement and manually measured displacement, 
\begin{equation} \label{eqn:perceptionCalib}
    D = \sigma d
\end{equation}
where $D$ is the manually measured displacement, $\sigma$ is the multiplication factor, and $d$ is the displacement from fiducial marker detection. 

To better estimate $\sigma$, we have an overdetermined system by sampling a number of displacement points along the linear shaft $\{d\}$,
\begin{equation} \label{eqn:overdetermined_system}
        \bm{D} = \sigma \bm{d}
\end{equation}
where $\bm{D} = [D_1, D_2, \cdots, D_N]^{T}$ is a vector of manually measured displacements, and $\bm{d} = [d_1, d_2, \cdots, d_N]^{T}$ is a vector of fiducial marker displacements. The Moore–Penrose inverse is applied to solve the overdetermined system, 
\begin{equation}  \label{eqn:perception_calibration}
    \sigma = \bm{D}  \bm{d}^T  [\bm{d}  \bm{d}^T]^{-1}
\end{equation}

We utilize the limit switches at the ends of the linear shaft (i.e., bed position calibration switches shown in Fig.~\ref{fig:top-down_Shakebot}) to facilitate the measurement of bed displacement, $\bm{D}$. To begin the perception system calibration, the bed moves to the left end until the left bed position calibration switch is triggered. At the same time, the fiducial marker detection at this position is recorded. The bed then moves to the right end, triggering the right calibration switch, and the fiducial marker detection is recorded again. We input the manually measured length between the two calibration switches. During the perception calibration, we have a position controller that rotates the motor shaft by one angular step given one PWM signal. As the bed moves from one end to the other, we count and record the motor steps, from which we calculate the bed displacement per angular step (translational step resolution). Then, the bed randomly moves along the linear shaft to sample data points, such as step counts (from the left calibration switch) and fiducial marker detections. Using the step counts and translation step resolution, we calculate the actual bed displacements. The samples of the actual bed displacements and fiducial marker displacements are used in Eq.~\ref{eqn:perception_calibration} to estimate $\sigma$. 

\subsection{Velocity Controller Calibration}
\label{sec:velocity_controller_calibration}
The velocity controller calibration aims to reduce the uncertainties from the toothed belt transmission. Because of the closed-loop stepper motor, we assume that the error between the actual angular velocity and the desired angular velocity of the stepper motor is diminished. Given a desired translational velocity of the bed, we calculate the corresponding desired angular velocity. However, the toothed pulley radius $r$ is close but not equal to the outside radius of the pulley. The uncertainty in $r$ comes from factors such as the tightness of the toothed belt, manufacturing imprecision, and tooth geometry. We correct for such uncertainty using a multiplicative factor, 
\begin{equation}
    \omega'(t) = \frac{\gamma v(t)}{r}
\end{equation}
where $\gamma$ is a constant that captures the first-order systematic uncertainty in the toothed belt transmission. The velocity controller calibration needs to estimate a $\gamma$ that minimizes the error between the desired and actual translational velocities. Because $\gamma$ is a factor that calibrates velocity, we denote $\gamma v(t)$ as calibrated velocity. Instead of directly measuring translational velocity in real time, we conduct half-cosine displacement motions (i.e., $t\in[0,1/(2f)]$ in Eq.~\ref{eqn:displacement}) and estimate $\gamma$ from displacement differences,
\begin{equation} \label{eqn:displacement_factor}
    D' = \gamma D
\end{equation}
where $D$ is the desired half-cosine displacement, and $D'$ is the actual half-cosine displacement measured from the fiducial marker detection. Similarly, we randomly sample the half-cosine displacement motions and use the Moore–Penrose inverse to estimate $\gamma$. Using displacement difference to estimate velocity difference has the advantage of easy implementation, because directly measuring translation velocity is challenging.  

We briefly demonstrate that $\gamma$ estimated from the displacements is the same multiplicative factor for velocity calibration. Based on Eq.~\ref{eqn:displacement_factor}, we have calibrated displacement function,
\begin{equation} 
    D'(t) = \gamma A - \gamma A \cos(2 \pi ft).
\end{equation}
Note that $\gamma$ here is estimated from the displacement differences described above. We take the derivative of $D'(t)$ and obtain the calibrated velocity function,
\begin{equation} \label{eqn:velocity_factor}
    v'(t) = \dot{D'}(t)= 2\pi \gamma  Af \sin(2 \pi ft).
\end{equation}
We substitute Eq.~\ref{eqn:velocity} into Eq.~\ref{eqn:velocity_factor},
\begin{equation}
     v'(t) = \gamma v(t).
\end{equation}
Therefore, $\gamma$ estimated from displacements is also the factor that calibrates velocity. 

\section{Experiments and Discussion}
\label{sec:experiment}
\begin{figure}[h]
\captionsetup[subfigure]{justification=centering}
\begin{subfigure}[t]{4.7cm}
    \includegraphics[width=\textwidth]{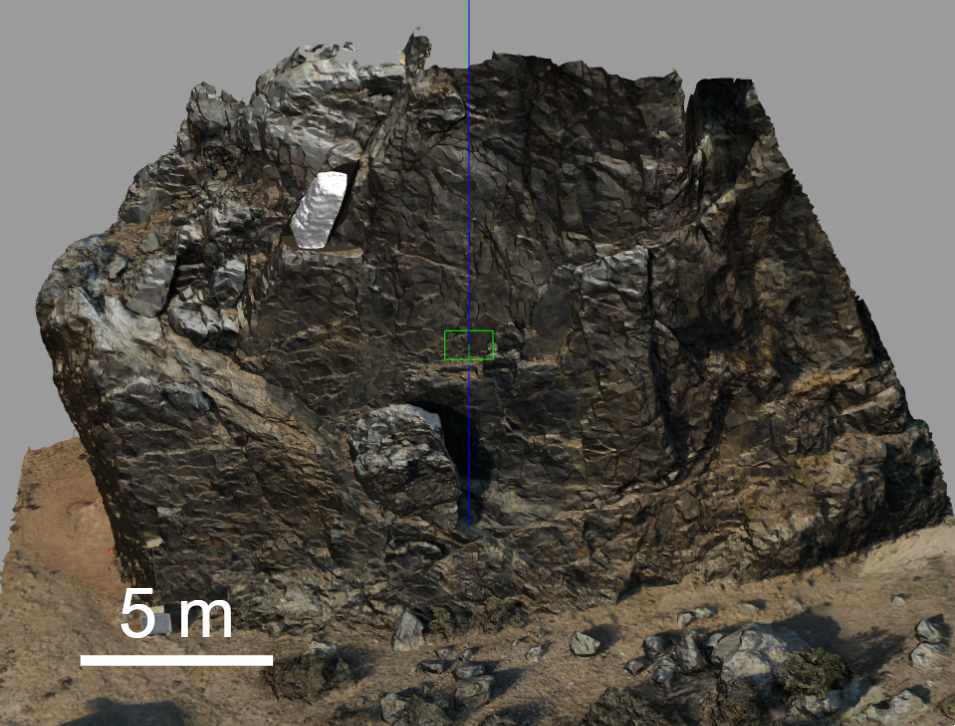}
    \caption{Double Rock site in coastal Central California}
    \label{fig:double_rock_site}
\end{subfigure}
\begin{subfigure}[t]{3.8cm}
    \centering
    \includegraphics[width=\textwidth]{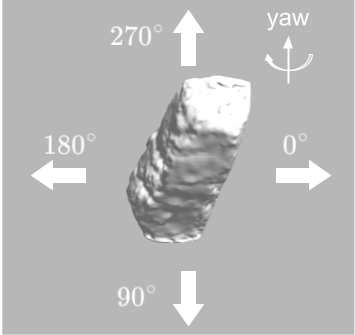}
    \caption{PBR extracted from Double Rock site}
    \label{fig:double_rock_orientation}
\end{subfigure}
\caption{Precariously Balanced Rock (PBR) at Double Rock site in coastal Central California. The yaw orientation indicates the relative direction of single-pulse cosine displacement motion.}
\label{fig:double_rock}
\end{figure}

We conducted experiments to demonstrate an application of the Shakebot in PBR fragility studies and to test its performance in validating simulation. These experiments focused on a PBR at the Double Rock site in coastal Central California \cite{rood2020earthquake}. The PBR, as shown in Fig.~\ref{fig:double_rock}, was mapped by an unpiloted aerial vehicle and constructed using structure from motion and Poisson reconstruction algorithms \cite{chen2024virtual}.  

\begin{figure}[h]
\vspace{8pt}
\captionsetup[subfigure]{justification=centering}
\begin{subfigure}[t]{4.2cm}
    \includegraphics[width=\textwidth]{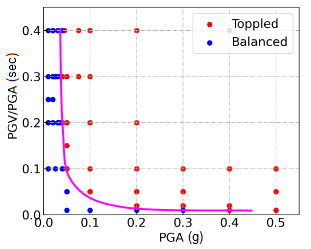}
    \caption{Yaw 0°}
    \label{fig:ExperimentResults_a}
\end{subfigure}
\begin{subfigure}[t]{4.2cm}
    \centering
    \includegraphics[width=\textwidth]{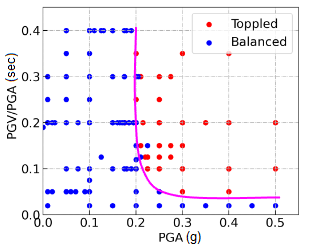}
    \caption{Yaw 270°}
    \label{fig:ExperimentResults_b}
\end{subfigure}
\caption{Overturning response diagrams of the Double Rock PBR from orientations of yaw 0° and yaw 270°. The red and blue dots represent overturning responses of being toppled and balanced after single-pulse cosine ground motions, respectively. The curves indicate the boundaries of the overturning responses.}
\label{fig:anisotropy_experiment_results}
\end{figure}

The first experiment demonstrates the Shakebot application in studying the fragility anisotropy of PBRs. We down-scaled the Double Rock PBR from 151.0 cm to 12.0 cm in height and 3D-printed it with PLA material. The resulting PBR, denoted as 3D-printed PBR(a), had a total mass of 105g. Because of the asymmetric geometry of 3D-printed PBR(a), the overturning responses were expected to vary with different ground motion directions. To investigate this fragility anisotropy idea, we placed the 3D-printed PBR(a) with two initial orientations (yaw angles of 0° and 270°, as shown in Fig.~\ref{fig:double_rock}) on the bed. Using the UI, we obtained the response diagrams from a set of ground motions, as shown in Fig.~\ref{fig:anisotropy_experiment_results}. The response diagrams indicated that 3D-printed PBR(a) oriented at a yaw angle of 0° was more fragile than at yaw angle of 270°. This result is consistent with previous studies showing that PBRs with smaller minimal contact angles along the motion direction are more fragile \cite{purvance2008freestanding, haddad2012estimating}. Addtionally, the resulting boundary curves closely resembled the simulation results of the Double Rock PBR with original shape \cite{chen2024virtual}. 

\begin{figure}[h]
    \centering
    \includegraphics[scale=0.4]{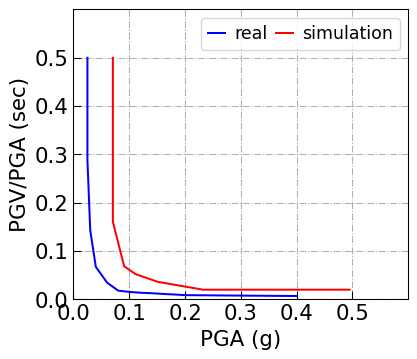}
    \caption{Overturning boundary curves from Shakebot and virtual shake robot. Reprinted from \cite{chen2022automated}. Reprinted with permission.}
    \label{fig:vsr_validation}
\end{figure}

In the second experiment, we used the Shakebot to evaluate the overturning simulation. The Double Rock PBR was down-scaled to 12.8 cm in height and 3D-printed with the PETG material. The resulting PBR, denoted as 3D-printed PBR(b), had a total mass of 404 g. We imported the 3D model of the down-scaled PBR to Gazebo and conducted overturning experiments using the virtual shake robot \cite{chen2024virtual}. To examine the simulation results, we used the Shakebot to overturn 3D-printed PBR(b). In both simulation and real-world experiments, the PBR had a yaw angle of 0°. As shown in Fig.~\ref{fig:vsr_validation}, the overturning boundary curves from the real-world and simulation experiments are similar. Future work should explore the causes of their difference. For example, we observed that Gazebo reduced the precision of configuration parameters when passing those parameters to the Bullet physics engine. The data from real-world experiments will aid in calibrating and improving the simulation experiments. 

\begin{figure}[h]
    \centering
    \includegraphics[width=0.48\textwidth]{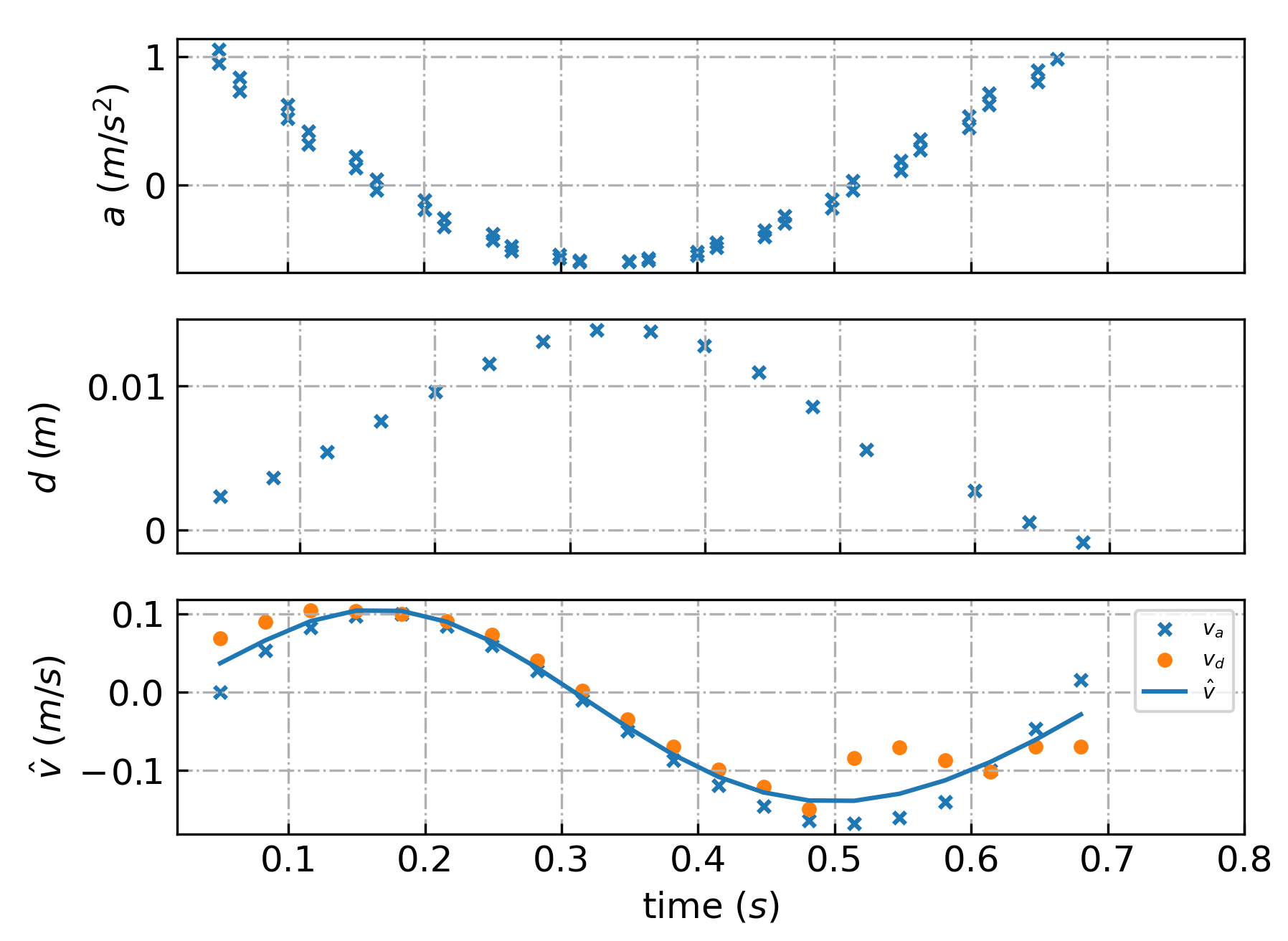}
    \caption{Velocity estimation from regression of acceleration and displacement.}
    \label{fig:velocity_fusion_results}
\end{figure}

Fig.~\ref{fig:velocity_fusion_results} shows the estimation results of acceleration, displacement, and velocity from the input PGA of 0.98 $m/s^2$ and PGV/PGA of 0.1 $s$. According to the PGA and PGV/PGA, the desired PGV and peak ground displacement are 0.1 $m/s$ and 0.01 $m$, respectively. The acceleration estimation and displacement estimation can be improved with better measurement equipment such as a better-quality accelerator and motion capture system. Our velocity estimation method used the 6\textsuperscript{rd} order polynomial function to fuse data from afford sensors, resulting in a PGV estimate of 0.12 $m/s$. The resulting velocity estimation is better than individual velocities derived from the accelerometer and fiducial marker detection. Future work should quantify uncertainties in displacement, velocity, and acceleration estimates. Investigations on other regression models (e.g., Gaussian process regression) and real-time state estimation may also improve the ground motion estimation. 

Because validating ground motion estimation requires future investment in expensive measurement equipment, we assumed that the actual ground motions were the same as the desired ground motions during the overturning experiment. This assumption is reasonable given our hierarchical control system and calibration processes. However, future work should verify this assumption. We also anticipate that the PBR response status registration can be automated by an object detection system and PBR placement can be implemented by a robotic arm. By automating the overturning experiments, we can save on human labor and ensure accurate PBR placement for reliable results.

\section{Conclusion}
\label{sec:conclusion}
We introduced a low-cost, open-source shake table named \textit{Shakebot} for earthquake research and education. Leveraging concepts and tools from robotics, we simplified and advanced the hardware and software compared to the existing shake tables. We presented the technical details of the perception and motion systems, which are important for the open-source community to upgrade or customize the Shakebot. Experiments showed the applications of the Shakebot for PBR studies. Our work demonstrates an interdisciplinary method of using robotics for natural sciences, providing a step towards robotics-enabled automated geoscience \cite{chen2022automated}. The Shakebot provides an affordable platform for education, outreach, and other ground motion experiments.

\section*{ACKNOWLEDGEMENTS}
This research was supported by the Pacific Gas and Electric Company and Southern California Earthquake Center (Contribution No. 21137). SCEC is funded by NSF Cooperative Agreement EAR-1600087 and USGS Cooperative Agreement G17AC00047. Thank you to Chris Madugo for his support and advice in the research. 


\printbibliography

@article{gavridou2017shake,
  title={Shake-table test of a full-scale 4-story precast concrete building. I: Overview and experimental results},
  author={Gavridou, Sofia and Wallace, John W and Nagae, Takuya and Matsumori, Taizo and Tahara, Kenich and Fukuyama, Kunio},
  journal={Journal of Structural Engineering},
  volume={143},
  number={6},
  pages={04017034},
  year={2017},
  publisher={American Society of Civil Engineers}
}

@article{elgamal2005line,
  title={On-line educational shake table experiments},
  author={Elgamal, Ahmed and Fraser, Michael and McMartin, Flora},
  journal={Journal of Professional Issues in Engineering Education and Practice},
  volume={131},
  number={1},
  pages={41--49},
  year={2005},
  publisher={American Society of Civil Engineers}
}

@article{rakicevic2021hybrid,
  title={A hybrid seismic isolation system toward more resilient structures: Shaking table experiment and fragility analysis},
  author={Rakicevic, Zoran and Bogdanovic, Aleksandra and Farsangi, Ehsan Noroozinejad and Sivandi-Pour, Abbas},
  journal={Journal of Building Engineering},
  volume={38},
  pages={102194},
  year={2021},
  publisher={Elsevier}
}

@article{bogdanovic2019shake,
  title={Shake table tests and numerical investigation of a resilient damping device for seismic response control of building structures},
  author={Bogdanovic, Aleksandra and Rakicevic, Zoran and Noroozinejad Farsangi, Ehsan},
  journal={Structural Control and Health Monitoring},
  volume={26},
  number={11},
  pages={e2443},
  year={2019},
  publisher={Wiley Online Library}
}

@article{yim_rocking_1980,
	title = {Rocking response of rigid blocks to earthquakes},
	volume = {8},
	issn = {1096-9845},
	pages = {565--587},
	number = {6},
	journal = {Earthquake Engineering \& Structural Dynamics},
	author = {Yim, Chik-Sing and Chopra, Anil K. and Penzien, Joseph},
	year = {1980},
	langid = {english},
}

@mastersthesis{malyuta2017mt,
  author = {Danylo Malyuta},
  title = {{Guidance, Navigation, Control and Mission Logic for Quadrotor Full-cycle Autonomy}},
  language = {english},
  type = {Master thesis},
  school = {Jet Propulsion Laboratory},
  address = {4800 Oak Grove Drive, Pasadena, CA 91109, USA},
  month = dec,
  year = {2017}
}

@inproceedings{Wang2016,
  author = {Wang, John and Olson, Edwin},
  booktitle = {2016 IEEE/RSJ International Conference on Intelligent Robots and Systems (IROS)},
  isbn = {978-1-5090-3762-9},
  month = {10},
  pages = {4193--4198},
  publisher = {IEEE},
  title = {{AprilTag 2: Efficient and robust fiducial detection}},
  year = {2016}
}

@incollection{ding2020challenges,
  title={Challenges of reinforcement learning},
  author={Ding, Zihan and Dong, Hao},
  booktitle={Deep Reinforcement Learning},
  pages={249--272},
  year={2020},
  publisher={Springer}
}

@inproceedings{zhao2020sim,
  title={Sim-to-real transfer in deep reinforcement learning for robotics: a survey},
  author={Zhao, Wenshuai and Queralta, Jorge Pe{\~n}a and Westerlund, Tomi},
  booktitle={2020 IEEE Symposium Series on Computational Intelligence (SSCI)},
  pages={737--744},
  year={2020},
  organization={IEEE}
}

@mastersthesis{kinay2006construction,
  title={Construction and control of a desktop earthquake simulator},
  author={K{\i}nay, G{\"o}k{\c{c}}e},
  year={2006},
  school={{\.I}zmir Institute of Technology}
}

@article{shao2014development,
  title={Development of a versatile hybrid testing system for seismic experimentation},
  author={Shao, X and Enyart, G},
  journal={Experimental Techniques},
  volume={38},
  number={6},
  pages={44--60},
  year={2014},
  publisher={Springer}
}

@phdthesis{chen2022automated,
  title={Automated Geoscience with Robotics and Machine Learning: A New Hammer of Rock Detection, Mapping, and Dynamics Analysis},
  author={Chen, Zhiang},
  year={2022},
  school={Arizona State University}
}

@article{chen2024virtual,
  title={Virtual Shake Robot: Simulating Dynamics of Precariously Balanced Rocks for Overturning and Large-displacement Processes},
  author={Chen, Zhiang and Arrowsmith, Ram{\'o}n and Das, Jnaneshwar and Wittich, Christine and Madugo, Chris and Kottke, Albert},
  journal={Seismica},
  volume={3},
  number={1},
  year={2024}
}

@article{purvance2008freestanding,
  title={Freestanding block overturning fragilities: Numerical simulation and experimental validation},
  author={Purvance, Matthew D and Anooshehpoor, Abdolrasool and Brune, James N},
  journal={Earthquake Engineering \& Structural Dynamics},
  volume={37},
  number={5},
  pages={791--808},
  year={2008},
  publisher={Wiley Online Library}
}

@article{anooshehpoor2004methodology,
  title={Methodology for obtaining constraints on ground motion from precariously balanced rocks},
  author={Anooshehpoor, Abdolrasool and Brune, James N and Zeng, Yuehua},
  journal={Bulletin of the Seismological Society of America},
  volume={94},
  number={1},
  pages={285--303},
  year={2004},
  publisher={Seismological Society of America}
}

@article{shi1996rocking,
  title={Rocking and overturning of precariously balanced rocks by earthquakes},
  author={Shi, Baoping and Anooshehpoor, Abdolrasool and Zeng, Yuehua and Brune, James N},
  journal={Bulletin of the Seismological Society of America},
  volume={86},
  number={5},
  pages={1364--1371},
  year={1996},
  publisher={The Seismological Society of America}
}

@article{saifullah2022seismic,
  title={Seismic response of two freestanding statue-pedestal systems during the 2014 South Napa Earthquake},
  author={Saifullah, M Khalid and Wittich, Christine E},
  journal={Journal of Earthquake Engineering},
  volume={26},
  number={10},
  pages={5086--5108},
  year={2022},
  publisher={Taylor \& Francis}
}

@article{rood2020earthquake,
  title={Earthquake hazard uncertainties improved using precariously balanced rocks},
  author={Rood, AH and Rood, DH and Stirling, MW and Madugo, CM and Abrahamson, NA and Wilcken, KM and Gonzalez, T and Kottke, A and Whittaker, AC and Page, WD and others},
  journal={AGU Advances},
  volume={1},
  number={4},
  pages={e2020AV000182},
  year={2020},
  publisher={Wiley Online Library}
}

@incollection{anderson2014precarious,
  title={Precarious rocks: Providing upper limits on past ground shaking from earthquakes},
  author={Anderson, John G and Biasi, Glenn P and Brune, James N},
  booktitle={Earthquake hazard, risk and disasters},
  pages={377--403},
  year={2014},
  publisher={Elsevier}
}

@article{brune1996precariously,
  title={Precariously balanced rocks and ground-motion maps for southern California},
  author={Brune, James N},
  journal={Bulletin of the Seismological Society of America},
  volume={86},
  number={1A},
  pages={43--54},
  year={1996},
  publisher={The Seismological Society of America}
}

@article{jacobsen1929vibration,
  title={Vibration research at Stanford university},
  author={Jacobsen, Lydik S},
  journal={Bulletin of the Seismological Society of America},
  volume={19},
  number={1},
  pages={1--27},
  year={1929},
  publisher={The Seismological Society of America}
}

@article{lu2008shake,
  title={Shake table model testing and its application},
  author={Lu, Xilin and Fu, Gongkang and Shi, Weixing and Lu, Wensheng},
  journal={The Structural Design of Tall and Special Buildings},
  volume={17},
  number={1},
  pages={181--201},
  year={2008},
  publisher={Wiley Online Library}
}

@article{haddad2012estimating,
  title={Estimating Two-dimensional Static Stabilities and Geomorphic Settings of Precariously Balanced Rocks from Unconstrained Digital Photographs},
  author={Haddad, David E and Zielke, Olaf and Arrowsmith, J Ram{\'o}n and Purvance, Matthew D and Haddad, Amanda G and Landgraf, Angela},
  journal={Geosphere},
  volume={8},
  number={5},
  pages={1042--1053},
  year={2012},
  publisher={Geological Society of America}
}

@misc{web:qanser_st2,
%     author = {Quanser},
%     title = {Quanser Shake Table II},
%     year = {1989},
%     url = {https://www.quanser.com/products/shake-table-ii/}
% }

@electronic{web:camera_calibration,
 author   = "ROS.org",
 title     = "camera\_calibration",
 url       = "http://wiki.ros.org/camera_calibration"
}

@electronic{web:testbox_shaketable,
 author   = "QUAKELOGIC",
 title     = "Testbox-shaketable",
 url       = "https://quakelogic.net/actuators-shaketables/testbox_shaketable"
}

@article{baran2011construction,
  title={CONSTRUCTION AND PERFORMANCE TEST OF A LOW-COST SHAKE TABLE},
  author={Baran, T and Tanrikulu, AK and Dundar, C and Tanrikulu, AH},
  journal={Experimental Techniques},
  volume={35},
  number={4},
  pages={8--16},
  year={2011},
  publisher={Blackwell Publishing Ltd Oxford, UK}
}

@misc{web:kelvin_es,
    author = {Kelvin Educational},
    title = {Kelvin advanced earthquake simulator},
    url = {https://kelvin.com/kelvin-e-q-machine-2021/}
}

@article{damci2019,
	Author = {Damcı, E. and {\c S}ekerci, {\c C}.},
	Da = {2019/04/01},
	Id = {Damcı2019},
	Journal = {Experimental Techniques},
	Number = {2},
	Pages = {179--198},
	Title = {Development of a Low-Cost Single-Axis Shake Table Based on Arduino},
	Ty = {JOUR},
	Volume = {43},
	Year = {2019},}

\end{document}